\documentclass[11pt]{article}

\usepackage{natbib}
\usepackage{moreverb}
\usepackage{rotating}

\usepackage{fullpage}
\usepackage{graphicx}

\usepackage{epsf,pstricks}
\usepackage{amsmath}
\usepackage{amsbsy}
\usepackage{amsfonts}
\usepackage{xcolor}
\usepackage{bbm}
\usepackage{multirow}
\usepackage{textcomp}
\usepackage{caption}
\usepackage{subcaption}
\usepackage{epstopdf}
\DeclareGraphicsExtensions{.eps}

\usepackage[
			pdfpagelabels,hypertexnames=true,
			plainpages=false,
			naturalnames=false]{hyperref}
\usepackage{color}
\definecolor{darkblue}{rgb}{0,0.1,0.5}
\hypersetup{colorlinks,
			linkcolor=blue,
			anchorcolor=blue,
			citecolor=blue}

\usepackage{pgf}
\usepackage{tikz}
\usetikzlibrary{shapes,arrows}

\newcommand\mbf[1]{\boldsymbol{#1}}
\newcommand\mrm[1]{\mathrm{#1}}
\newcommand\indicatorBig[1]{{\mathbb{I}}[#1]}
\newcommand\indicator[1]{{\mathbb{I}}(#1)}

\title{ Data mining for censored time-to-event data: A Bayesian network model for predicting cardiovascular risk from electronic health record data }

\author{ Sunayan Bandyopadhyay\footnotemark[1], Julian Wolfson\footnotemark[2], David M. Vock\footnotemark[2], \\
Gabriela Vazquez-Benitez\footnotemark[3], Gediminas Adomavicius\footnotemark[4], Mohamed Elidrisi\footnotemark[1], \\
Paul E. Johnson\footnotemark[4], and Patrick J. O'Connor\footnotemark[3] }

\begin{document}
\maketitle

\footnotetext[1]{Department of Computer Science and Engineering, University of Minnesota, Minneapolis, MN}
\footnotetext[2]{Division of Biostatistics, University of Minnesota, Minneapolis, MN. Correspondence to: julianw@umn.edu}
\footnotetext[3]{HealthPartners Institute for Education and Research, Minneapolis, MN}
\footnotetext[4]{Department of Information and Decision Sciences, Carlson School of Management, University of Minnesota, Minneapolis, MN}

\begin{abstract}
Models for predicting the risk of cardiovascular events based on individual patient characteristics are important tools for managing patient care. Most current and commonly used risk prediction models have been built from carefully selected epidemiological cohorts. However, the homogeneity and limited size of such cohorts restricts the predictive power and generalizability of these risk models to other populations. Electronic health data (EHD) from large health care systems provide access to data on large, heterogeneous, and contemporaneous patient populations.  The unique features and challenges of EHD, including missing risk factor information, non-linear relationships between risk factors and cardiovascular event outcomes, and differing effects from different patient subgroups, demand novel machine learning approaches to risk model development. In this paper, we present a machine learning approach based on Bayesian networks trained on EHD to predict the probability of having a cardiovascular event within five years. In such data, event status may be unknown for some individuals as the event time is right-censored due to disenrollment and incomplete follow-up. Since many traditional data mining methods are not well-suited for such data, we describe how to modify both modelling and assessment techniques to account for censored observation times. We show that our approach can lead to better predictive performance than the Cox proportional hazards model (i.e., a regression-based approach commonly used for censored, time-to-event data) or a Bayesian network with {\em{ad hoc}} approaches to right-censoring. Our techniques are motivated by and illustrated on data from a large U.S. Midwestern health care system.
\end{abstract}


\section{Introduction}
In the United States, myocardial infarctions (MI) and strokes are the first and fourth leading causes of death and, in addition to contributing significantly to mortality, these conditions account for substantial morbidity with treatment costing more than \$300 billion annually \citep{Go_2014}. Clinical risk prediction scores or algorithms remain important tools for managing patient care and improving outcomes in the population. Broadly speaking, risk prediction scores can raise awareness of the substantial burden of cardiovascular disease (CVD) and risk factors associated with developing CVD \citep{LloydJones_2010}. In the clinical setting, accurate personalized cardiovascular risk prediction may identify patients at high risk for experiencing cardiovascular (CV) events (e.g., MI, stroke) so that clinicians may develop an appropriate intervention strategy and patients are motivated to remain adherent to that strategy.

Recent systematic reviews found that there are over 100 risk models produced between 1999 and 2009 \citep{Cooney_2009,Cooney_2010,matheny2011systematic} including the well-known Framingham \citep{DAgostino_2008}, SCORE \citep{Conroy_2003}, ASSIGN-SCORE \citep{Woodward_2007}, QRISK1 \citep{HippisleyCox_2007,HippisleyCox_2008}, QRISK2 \citep{HippisleyCox_2008a}, PROCAM \citep{Assmann_2002}, WHO/ISH, and Reynolds Risk Score \citep{Ridker_2007,Ridker_2008}. Although there are many risk prediction models for cardiovascular disease, nearly all have been estimated using data from carefully selected epidemiological cohorts. For example, the Framingham risk score is trained on a data set which excludes patients that have had a previous CV event, which represents a predominantly Caucasian population, and which includes patients from the late 1960s \citep{DAgostino_2008}. As a result of estimating the risk of CV events using data from these homogeneous cohorts, existing risk models are likely to only give accurate predictions for patients who are well represented in the training data sets. \cite{Collins_2009} provide an excellent illustration of the poor performance of the Framingham risk equations when applied to a contemporary population in the United Kingdom. Models constructed from a more diverse cohort are likely to produce more accurate estimates of CVD risk for a wider range of patients seen in the primary care clinic.

One source of clinical data is electronic health data (EHD) collected by from a health maintenance organization (HMO).  These data consist of electronic medical records (EMRs), insurance claims data, and mortality data obtained from the state government. EHD are increasingly available within the context of large health care systems and capture the characteristics of a heterogeneous population receiving care in a contemporary clinical setting. EHD databases typically include records on hundreds of thousands to millions of individual patients; therefore, a risk prediction model constructed from EHD could yield more accurate and generalizable risk predictions because even relatively specific sub-populations (e.g., patients with multiple comorbidities) are likely to be well-represented in such a large database.

\subsection{Challenges of electronic health data}

Although the scale and complexity of EHD provide great opportunities to improve cardiovascular risk prediction for contemporary patient populations, these data also pose serious challenges to those seeking to mine it, especially in comparison to data collected from epidemiological cohorts. Here, we briefly describe some of these challenges, all of which are present in the data set we analyze in Section \ref{sect:data-analysis}.

{\textbf{Missing data:}} EMR data are purely observational in the sense that they record individuals' encounters with the health care system, however regular or irregular those encounters may be. Without a structured schedule for obtaining lab measurements, these measurements may be obtained sporadically or simply be unavailable. Further, even when individuals do make regular clinical visits, clinical practice guidelines may recommend against obtaining certain measurements within particularly low-risk populations. For instance, it is uncommon for individuals under the age of 40 to obtain cholesterol measurements as part of a routine checkup. Several statistical techniques exist for imputing missing data, but the scale and complexity of EMR data make many of these unappealing.

{\textbf{Subgroup heterogeneity:}} EHD comprehensively describe diverse, contemporaneous patient populations. In contrast to a cohort study, patients do not have to satisfy eligibility criteria to appear in the database. Though in some cases it may make sense to restrict an EHD-based study to a narrowly defined sub-population (e.g., individuals with diabetes), our goal is to develop methods which make maximal use of the available data. Hence, the advanced risk prediction methods should be capable of using the data to automatically identify non-linear relationships between risk factors and CV events as well as sub-populations where the effects of those factors differ.

{\textbf{Sample size:}} The enormous sample size afforded by EHD is simultaneously a blessing and a curse. While there are ample data to support flexible nonparametric risk models, the complexity of these models is restricted by the ability to train them on hundreds of thousands or millions of records in a reasonable amount of time.

{\textbf{Incomplete follow-up and censoring:}} Our motivating data set is drawn from 10 years of EHD derived from an HMO. However, not all subjects were enrolled in the HMO for the full 10-year period, either because their first enrollment came after the beginning of this period or because they terminated insurance coverage (e.g., due to a change in employer) during the period. In this paper, we seek to estimate the 5-year risk of cardiovascular events. But the occurrence of cardiovascular events is not recorded in the EMR or in claims data during times when subjects are not enrolled, and a substantial fraction of subjects (71\% in our data) do not have enough follow-up data in the EMR to ascertain if they experienced a cardiovascular event over a 5-year period. In the language of statistical survival analysis, such subjects are said to be \emph{right-censored}.

The standard statistical methods to model the relationship between risk factors and time-to-event outcomes are the Cox proportional hazards model \citep{Cox_1972} and, to a lesser extent, the accelerated failure time model \citep{Buckley_1979}. Although these methods handle censored outcomes, they do not natively address the first three challenges encountered in EHD. In particular, the proportional hazards model assumes that the risk factors have a linear relationship with the log hazard of experiencing a CV event. If the analyst has {\em{a priori}} knowledge that this relationship is non-linear or differs in certain sub-groups, she may include non-linear transformations of  predictors or interactions between predictors, but this is often based on trial and error. Secondly, standard software for proportional hazards models removes subjects with any missing features which is a large percentage for EHD, and, as noted above, imputation methods may be onerous given the scale and complexity of EHD. Finally, as noted by \citep{Kattan_1998}, the proportional hazards model will show improved fit when additional terms are added in the model and there is no reliable indication when the model has been over fit.

The first three challenges (missing data, subgroup heterogeneity, and data dimensionality) motivate the use of a sophisticated, flexible, and scalable machine learning technique such as a Bayesian Network to mine EHD. However, the fourth challenge (incomplete follow-up) is not directly addressed by usual typical machine learning approaches. In the next section, we discuss the implications of censoring on the application of machine learning techniques.

\subsection{Machine learning with censored outcome data}

Many data mining techniques have been developed for the related tasks of classification and class probability estimation for a binary outcome. These methods have been shown to yield a performance that is superior to traditional regression-based techniques in healthcare-related classification problems ~\citep{song2004comparison,colombet2000models}, and are more adept at handling the aforementioned characteristics of EHD, including correlation and non-linear relationships between the features. However, they are not designed for use in circumstances where outcomes may be censored.

Fully supervised machine learning methods assume that the outcome is known for all subjects, but in our setting the binary outcome (whether or not a patient experiences a cardiovascular event within 5 years) is undetermined for subjects who are censored, i.e., who do not experience an event but do not have a full 5 years of follow-up. Simplistic techniques to deal with this issue, such as discarding censored observations (see, e.g., \cite{Larranaga_1997,Sierra_1998,Blanco_2005}) or treating them as zeroes (non-events), are known to induce bias in the estimation of class probabilities \citep{Kattan_1998}, making typical fully supervised classification approaches unsuitable. For example, \cite{Stajduhar_2009} demonstrated the impact of unaccounted-for censoring on the construction and performance of Bayesian networks. Semi-supervised approaches are also generally not applicable since the labeled (non-censored) and unlabeled (censored) observations are not samples from the same underlying population. Furthermore, censored observations are not truly `unlabeled' since they carry useful partial information about the outcome (i.e., that the event outcome did not occur before the subject was censored).

There has been an increasing interest to adapt machine learning techniques to this type of censored, time-to-event data \citep{Lucas_2004}. As one example, \cite{ishwaran2008random} make use of specialized random survival forests to analyze right-censored survival data, where a survival score is associated with every terminal node of the trees in the forest.  Inspired by the growing need and opportunities for machine learning approaches to complex healthcare data-driven problems, we propose a general-purpose extension of Bayesian networks using inverse probability of censoring weights (IPCW). The resulting technique properly accounts for censoring while retaining the features which make this flexible machine learning approach appealing in the context of EHD. We also show that traditional evaluation metrics for classification accuracy (e.g., receiver-operating curve, net reclassification improvement) may be misleading in the presence of censoring, and describe better alternatives.

\section{Bayesian networks for electronic health data}\label{sec:BNforEHR}

\subsection{Why Bayesian networks?}

Bayesian networks are especially well-suited to handle the intricacies of risk prediction using EHD. Compared to support vector machines or neural networks, Bayesian networks have a clear edge in interpretability, which is important to the end-users of these prediction models in the healthcare domain (e.g., physicians and clinical researchers). The Bayesian network framework we adopt handles missing data naturally and efficiently, eliminating the need for electronic health data sets to be ``pre-imputed'' to produce complete data, and is also computationally efficient.

Because of their interpretability and their ability to aid in reasoning with uncertainty, Bayesian networks have been used extensively in biomedical applications (see \cite{Lucas_2004} for a review). In particular, they have been used to aid in understanding of disease prognosis and clinical prediction \citep{Andreassen_1999,Verduijn_2007,Lipsky_2005,Sarkar_2013,Frances_2013,Lappenschaar_2013}; used to guide the selection of the appropriate treatment \citep{Lucas_2000,Kazmierska_2008,Smith_2009,Yet_2013,Velika_2014}; and implemented as part of clinical decision support systems \citep{Lucas_1998,Sesen_2013}.

In spite of the wide applicability in biomedical applications, there is a limited amount of previous work on the application of Bayesian networks to censored outcome data \citep{Lucas_2004}. \cite{Zupan_2000} and \cite{Stajduhar_2010} have proposed approaches in which censored observations are repeated twice in the dataset, one as experiencing the event and one event-free. Each of these observations are assigned a weight based on the {\em{marginal}} probability of experiencing an event between the censoring time and $\tau$, the time the event status will be assessed. This approach, although intuitive, is provably biased and inconsistent because the method to weight each of the replicated observations is based on the marginal probability and does not properly account for the covariates. \cite{Stajduhar_2012} adopt a more principled likelihood-based approach to imputing event times, but their imputation technique may perform poorly if the assumed parametric distribution of event times is incorrect. Other approaches, including replacing the time-to-event with the martingale from the null model,  have been proposed to handle censored data in other machine learning methods including support vector regression, recursive partitioning, and multiple adaptive regression splines \citep{Therneau_1990,Kattan_1998,Kattan_2003}. However, these approaches require that the technique to mine the data permit a continuous outcome and are, therefore, not amenable to the Bayesian network approach considered here.

In the following, we develop an extension of Bayesian networks which accounts for right-censored event indicators using inverse probability of censoring weights (IPCW) \citep{Robins_2000,Bang_2000,Bang_2002,Rotnitzky_2005,Tsiatis_2006}.  We begin by reviewing a classical Bayesian network approach, operating on a binary indicator of whether or not each subject experienced an event during the follow-up period. Clinical knowledge is used to construct a graphical model relating the relevant risk factors to the probability of an event. We then describe a model-averaging strategy to control model complexity and stabilize risk predictions for small patient subgroups. The extension of IPCW to Bayesian networks is discussed before we describe how to extend traditional predictive performance evaluation metrics when outcomes may be right-censored.

\subsection{Bayesian networks}
\label{subsub:Bayes_Analysis}
Let the features recorded on a patient be represented by a $p$-dimensional vector $\mathbf{X} = (X_1 \cdots X_p)$ where $X_i$ is the $i^\mathrm{th}$ risk factor (some of the factors could be missing for certain patients). Let $E=1$ indicate that an event (e.g., a CV event) occurred for a given patient within $\tau$ years of the beginning of the follow-up period, and $E=0$ indicate the absence of such an event in that time frame. Though our ultimate goal is to handle the case where $E$ is unknown for some patients, for now we assume that at least $\tau$ years of follow-up is available on each patient so that $E$ is fully observed.

The target of estimation is $P(E=1|\mathbf{X})$; using Bayes theorem, one can rewrite this probability as

\begin{equation} \label{eq:bayesBasic}
P(E=1|\mathbf{X}) = \frac{P(\mathbf{X}|E=1)P(E=1)}{\sum_{e\in\{0,1\}} P(\mathbf{X}|E=e)P(E=e)},
\end{equation}
so that the focus is shifted to estimation of $P(\mathbf{X}|E=e)$ and $P(E=e)$ for $e = 0,1$. In general, dimensionality $p$ may be too large to make joint modeling of $P(\mathbf{X} | E=1)$ feasible.  To simplify the joint modeling task, one can represent the joint distributions of $\mathbf{X}|E=e$ using a directed acyclic graph (DAG), i.e., a Bayesian network. The DAG encodes conditional independence relationships between variables, allowing the joint distribution to be decomposed into a product of individual terms conditioned on their parent variables \citep{stuart2003artificial}:

\begin{equation} \label{eq:DAGfactorsCond}
P(\mbf{X}|E) = \prod_{i=1}^p P(X_i|\mrm{Pa}(X_i),E)
\end{equation}
where $\mrm{Pa}(X_i)$ are the parents of $X_i$.

\begin{figure*}[ht]
\centering
\includegraphics[width=0.75\textwidth]{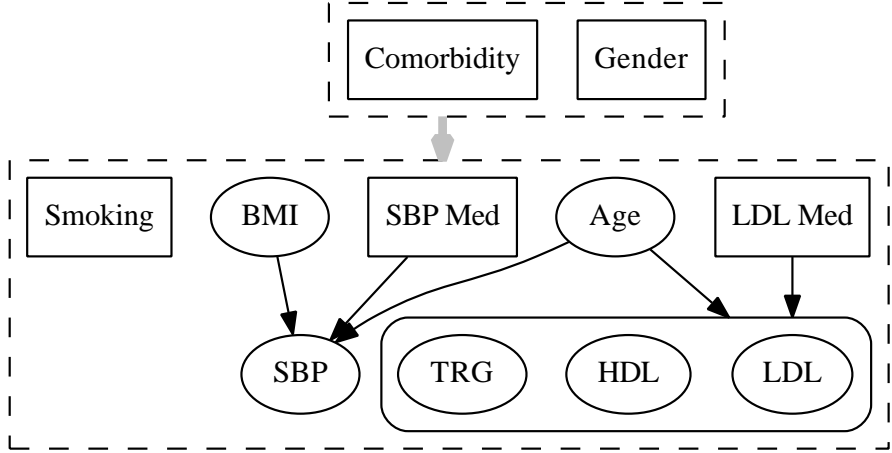}
\caption{The graphical model for our Bayesian network for CV risk prediction.  The figure includes the structure of risk factors, conditioned on the CV event status.  In particular, nodes represent input variables and edges represent conditional dependencies between the variables.  The reader should also assume that our outcome variable (CV Event) is connected to every node in the graph (omitted in the picture for parsimony).  Continuous and discrete variables are indicated by elliptical and rectangular nodes, respectively. Nodes in boxes with rounded corners indicate that they are modelled jointly. The nodes are grouped into subgraphs indicated by the dashed boxes. The grey edge between subgraphs indicates an edge from every node in the source subgraph to every node in the destination subgraph or node.  The full description of each of the features appears in Section~\ref{sec:data}. {\em Comorbidity}: Whether or not patient has pre-existing CVD; {\em Smoke}: current smoking status of patient; {\em BMI}: body mass index of patient; {\em SBP}: systolic blood pressure; {\em SBP Med}: Number of blood pressure medication classes currently prescribed to patient; {\em TRG}: triglycerides; {\em HDL}: high density lipoprotein; {\em LDL}: low density lipoprotein; (TRG, HDL, and LDL are components of patient's cholesterol measurement and are modelled jointly); {\em LDL Med}: indicator for whether the patient is on LDL lowering medication.}
\label{figure:BayesNet}
\end{figure*}

One advantage of the Bayesian network approach is that clinical knowledge can be used to suggest and refine DAG structures. While methods exist to infer the DAG structure from data, in our application we used parsimonious DAGs based on information from the medical literature as well as clinical judgment from the medical experts who collaborated on this research project. The DAG structure used in our predictive models is shown in Figure \ref{figure:BayesNet} (see the caption of the figure for a brief description) and is explained in greater detail in Section \ref{sec:data}. The graphical model in Figure \ref{figure:BayesNet} contains both continuous-valued nodes (which are elliptical in the figure) and discrete-valued nodes (which are rectangular). The network is therefore a hybrid Bayesian network~\citep{murphy1998inference}.

\subsection{Learning Bayesian network parameters}\label{sub:LNR}

To evaluate the joint distributions $P(\mbf{X}|E=e)$, for each feature $X_i$ (or group of features modelled jointly such as the lipid measures in our DAG) we construct vectors $\mbf{G}_i$ such that $\mbf{G}_i = (X_i,\mathrm{Pa}(X_i))$  and learn the joint distributions $P(\mbf{G}_i|E)$ of each of the groups. The conditional distributions $P(X_i | \mathrm{Pa}(X_i),E)$ of Equation \eqref{eq:DAGfactorsCond} can be derived from these joint distributions.  This approach differs slightly from the conventional strategy where a regression model is constructed to link the values of the parent nodes of $X_i$ to the mean of $X_i$.  Typically, we would assume that the mean of $X_i$ is linearly related to the values of the parent nodes, and we must learn or estimate the parameters in this linear regression model. Our approach provides more modeling flexibility (because we do not have to specify a regression model) without a substantial increase in computation due to our DAG structure where continuous nodes are arranged in relatively small, mutually independent groupings.

When all the components in $\mbf{G}_i$ are discrete variables, then evaluating the probability distributions $P(\mbf{G}_i|E)$ is simply a matter of estimating the frequency of each unique state of $\mbf{G}_i$ for $E=0$ or $E=1$.  $P(\mbf{G}_i|E)$ is represented as a probability distribution table of length $2 \times |\mbf{G}_i|$, where $|\mbf{G}_i|$ is the number of possible unique states of $\mbf{G}_i$.

Let $C_{G_i}$ be the dimension of $G_i$. When the $C_{G_i}$ components  in $\mbf{G}_i$ are continuous, the joint distribution $f_{\mbf{G}_i}$ (given $E=e$) is modelled as a mixture of $M$ multivariate normal distributions, as follows:

\begin{equation}\label{eq:JoinDistComps}
f_{\mbf{G}_i}(g|E=e)=\sum_{m=1}^M \rho_{m,e} \phi \left( \Sigma^{-1}_{m,e} ( g - \mu_{m,e}) \right),
\end{equation}
where $\phi$ is the density function of a $C_{G_i}$-variate standard normal random variable, $\mu_{m,e}$ and $\Sigma_{m,e}$ are the mean vector and variance matrix of $\mbf{G}_i$, and $\rho_{m,e}$ are the mixing parameters. The number of parameters to estimate is $2 \times M \times  C_{G_i} ( 1 + (C_{G_i} + 1)/2 )$.

When $\mbf{G}_i$ contains discrete parents (discrete children are not considered in our application), the components in $\mbf{G}_i$ are partitioned into $(\mathbf{Y}_i, \mathbf{Z}_i)$ where $\mbf{Y}_i$ represents the $C_{Y_i}$ continuous risk factors and $\mbf{Z}_i$ the $C_{Z_i}$ discrete risk factors. The continuous components $\mathbf{Y}_i$ are then estimated using the same mixture approach as in Equation~\eqref{eq:JoinDistComps}, conditional on the levels of $\mathbf{Z}_i$:

\begin{equation}\label{eq:JoinDistComps2}
f_{\mbf{Y}_i}(y|\mbf{Z}_i=z,E=e)=\sum_{m=1}^M \rho_{m,e,z} \phi \left( \Sigma^{-1}_{m,e,z} ( y - \mu_{m,e,z}) \right).
\end{equation}

If each component $Z^k_i$ of $\mathbf{Z_i}$ has $Q_i^k$ possible states, then $|\mbf{Z}_i| = \prod_{k=1}^{p_{z_i}}Q_i^k$. The resulting number of parameters to estimate is $2 \times M \times  |\mbf{Z}_i| \times C_{Y_i} ( 1 + (C_{Y_i} + 1)/2 )$.

For a fixed number of mixing components $M$, a standard expectation maximization (EM) algorithm is used to solve for the maximum likelihood estimators of the mean, variance, and mixing parameters.  In general, in order to control overfitting, one could consider $M$ to be a tunable parameter and select the number of mixture components using the Bayes Information Criteria (BIC) or some other goodness-of-fit measure.  In the following section, we describe a model averaging procedure to combine results across multiple values of $M$.

\subsection{Model averaging}\label{sect:MA}
Increasing the number of mixture components in our models can lead to overfitting the training data. This is even more likely when we try to model distributions that are represented by small subgroups of patients. We can prevent overfitting by evaluating the weighted average of the predictions of multiple models of varying complexity trained on samples drawn from the data. The weights are derived from the BIC, which penalizes complexity of the model, and are approximations to the posterior probability of the model \citep{Jackson_2009}. Although model averaging has been used to average over different network structures in Bayesian networks \citep{Tian_2010}, to the best of our knowledge it has not been previously used to average over models of varying complexity within a single network structure.

To implement the model averaging, we take $T$ bootstrap samples from our training data. For each bootstrap sample, we fit models with $q = 1, \dots, Q$ mixture components.  In our analysis of cardiovascular risk, we take $Q=4$. Generally, $Q$ should be large enough so that the largest values of $Q$ do not receive much weight. The weights for each level of model complexity $q$ are computed as follows:

\begin{equation} \label{eq:model_avg_weight}
w_q = \frac{\exp(-0.5 B_q)}{\sum_Q \exp(-0.5 B_q)},
\end{equation}
where $B_q= \frac{1}{T} \sum_{j=1}^T B_{q,j}$ and  $B_{q,j}$ is the BIC of the model fit on the $j^{th}$ bootstrap sample with $q$ mixtures. The BIC is computed as $-2\log(\mathbf{L}) + k\log(n)$ where $\mbf{L}$ is the log likelihood evaluated at the maximum likelihood parameter estimates, $n$ is the number of subjects included in the training set, and $k$ the number of (free) parameters included in the model.

The probability distribution of an ensemble is simply evaluated by finding the weighted mean of the probability of each mixture model in the ensemble:

\begin{equation}\label{eq:wtdPDF1}
P( \mbf{X} | E; \mbf{\theta},\mbf{\rho}) = \sum_{1\leq q \leq Q}w_q\left(\frac{1}{T}\sum_{1\leq t \leq T}\left(\sum_{1\leq i \leq q}\rho_i P(\mbf{X}|E;\mbf{\theta_{q,t,i}})\right)\right),
\end{equation}

\noindent where $\theta_{q,t,i}$ is the set of model parameters to be estimated.

\subsection{Extension to censored time-to-event data}\label{sect:IPCW}
Our development thus far has assumed that the presence or absence of a CV event $E$ is fully observed for all patients in the data set, but this is unlikely to be true when using information from real-world EHD. In this section, we show how it is possible to use a Bayesian network to predict the risk of a CV event in $\tau$ years when the event status at $\tau$ years cannot be assessed in some patients due to right-censoring. More specifically, in our application, once a patient leaves the health system or the study ends, their health state (i.e., risk factors - covariates of the risk prediction model) and event history are no longer recorded in the EMR. That is, the patient's follow-up ends when the subject leaves the health system.  Define $T$ as the time between the beginning of the follow-up period and a CV event, and define $C$ as the time between the beginning of the follow-up period and disenrollment or the end of the data capture period. We observe $V=\min(T,C)$ and $\delta=\indicator{T<C}$, the indicator for whether or not a CV event occurs.  If $\delta = 0$, the subject's event time is said to be censored. We can only ascertain the value of $E$ if $\delta=1$, or $\delta=0$ and $V > \tau$.

As mentioned earlier, one naive approach to handle these subjects for whom we cannot ascertain the value of $E$ would be to exclude them from our training data set or to set $E=0$, but both approaches would lead to biased estimators of the CV risk.  Instead we propose to adjust for right-censoring using an inverse probability of censoring weighting (IPCW) approach. This approach to handling censored event times assuming that the censoring time $C$ is independent of the CV event time $T$ and all features $\mbf{X}$. In our study, most patients are censored due to the end of the study or because they disenroll from the HMO due to a change in employment. The assumption is reasonable because the follow-up for most patients ends for reasons unrelated cardiovascular health. Let $G(t) = P(C>t)$ be the probability that the censoring time is greater than $t$.  We can estimate $G(t)=P(C>t)$, using the Kaplan-Meier estimator of the survival distribution (i.e., 1 minus the cumulative distribution function) of the censoring times.  The Kaplan-Meier estimator \citep{Kalbfleisch_2002} of the censoring process is given by

\begin{equation}
\hat{G}(t)=\prod_{i:t_{i}<t}\left( \frac{n_{i}-d^*_{i}}{n_{i}} \right ),
\end{equation}
where $d^*_{i}$ are the number of subjects who were censored at time $t_{i}$ and $n_{i}$ are the number of subjects ``at risk'' (i.e., number of subjects not previously censored or experiencing a CV event)  at time $t_{i}$. Unlike other {\em{ad hoc}} approaches to handling censored observations, the Kaplan-Meier (or product-limit) estimator is consistent (i.e., asymptotically unbiased) and efficient even in the presence of censored observations \citep{Kalbfleisch_2002}.

Then, for each patient $i$, we define a weight
\[
\omega_i = \left\{ \begin{array}{cl}
	0 & \text{if } \delta_i=0 \text{ and } V_i < \tau\\
	\frac{1}{ \hat{G}( \min(V_i, \tau) ) } & \text{otherwise} \end{array} \right.
\]
To fit the Bayesian network using IPCW, estimation of the parameters in \eqref{eq:wtdPDF1} is carried out using a weighted version of the EM algorithm described above, where the contribution of the $i^{th}$ subject to the likelihood is weighted by $\omega_i$. Details of the weighted EM algorithm are provided in \citet{fraley2012mclust}.  The weights are computed on the full training sample, and are not re-estimated for each bootstrap sample. Many software packages implementing the EM algorithm (e.g., Matlab, R) allow weights to be provided as arguments to the EM function, making the IPCW Bayesian network approach straightforward to implement.

In the IPCW approach, only those patients for whom we can determine $E$ contribute to the analysis, but they are reweighted to accurately ``represent'' the patients who were censored prior to $\tau$ and were, therefore, omitted from the analysis. For example, patients that have a longer time-to-event are more likely to be censored ($G$ is smaller) and hence receive larger weights. Note that for subjects with $E=0$ (and $V > \tau$) the weights for all individuals are $1/\hat{G}(\tau)$ so the maximum likelihood estimators for $P(\mbf{G}_i|E_i=0)$ are the same as in the unweighted analysis.

\subsection{Obtaining predictions}

To obtain the prediction $P(E=1|\mbf{X})$, we need to evaluate the right-hand side of Equation~\eqref{eq:bayesBasic}, where $P(\mbf{X}|E)$ is given by the product in Equation~\eqref{eq:DAGfactorsCond}. As noted above, the joint modeling described in Section \ref{sub:LNR} yields conditional distributions via the expression
\begin{align}
P(X_i|\mrm{Pa}(X_i),E) &= P(X_i|\mbf{G}_i \setminus X_i, E) \nonumber \\
	& = \frac{ P( \mbf{G}_i | E) }{ P(\mbf{G}_i \setminus X_i | E) } \label{eq:CondTermGeneral},
\end{align}
where $\mbf{G}_i \setminus X_i$ are the components of the vector $\mbf{G}_i$ not included in $X_i$. When $X_i$ is discrete (implying that $\mbf{G}_i$ contains only discrete nodes), the terms in Equation~\eqref{eq:CondTermGeneral} are derived from a probability table summarizing the distribution of $\mbf{G}_i$, given $E$. When $X_i$ is continuous, the densities are computed by plugging the estimated mean, variance matrix, and mixing parameters into Equation~\eqref{eq:wtdPDF1}.

\subsubsection{Missing features}

Within EHD, it is relatively common for attributes to be unmeasured on certain patients; for example, cholesterol measurements are typically not available on patients under the age of 40. If $X_i$ is missing, then the corresponding product term from Equation~(\ref{eq:bayesBasic}) is dropped. When any attribute from $\mathrm{Pa}(X_i)$ is missing (say $A_i$), we implicitly marginalize over $A_i$ by computing $P(\mbf{G}_i \setminus A_i |E) = \int_{a} P( \mbf{G}_i | A_i=a,E) dP_{A_i}(a|E)$ where $P_{A_i}(a|E)$ is estimated from subjects with $A_i$ observed.  In our data, information regarding medications, gender, and comorbidity is never missing; therefore, we are concerned only about marginalizing the continuous part of the distribution.

\subsection{Performance evaluation metrics} \label{sce:performanceMetrics}

The challenge of machine learning for censored data extends beyond improving existing methods to handle censoring. As we discuss in this section, the usual performance metrics applied to classification and prediction problems can be misleading when outcomes are subject to censoring. Here, we present generalizations of standard calibration (goodness-of-fit) and discrimination (concordance-index and net reclassification improvement) metrics which properly account for censored data and allow model performance to be assessed more accurately.

\subsubsection{Calibration}\label{sec:calibration}

For standard binary classification problems, calibration is commonly assessed by ranking the predicted class probabilities for the test set, binning the ranked predictions (e.g., by decile), and comparing the mean (or median) predicted class probability in each bin to the empirical class probability of the instances in that bin. We adopt a similar approach here, except that the empirical probability of experiencing an event prior to time $\tau$ within each bin is computed via the Kaplan-Meier estimator to properly account for censoring. Calibration plots compare predicted and Kaplan-Meier probabilities of experiencing an event before $\tau$ within bins defined by ranges of predicted probabilities. We also compute a calibration statistic,
\[
K = \sum_{j=1}^B \frac{ ( \bar p_j - p^{KM}_j )^2 }{ \bar p_j (1 - \bar p_j) }
\]
where $B$ is the number of bins, $\bar p_j$ is the average of predicted probabilities in bin $j$, and $p^{KM}_j$ is the Kaplan-Meier estimate of experiencing an event before $t$, computed form the data in bin $j$. $K$ is analogous to the $\chi^2$ statistic for assessing the calibration of logistic models suggested by \citep{Hosmer_1980,Lemeshow_1982}.

\subsubsection{Concordance index}\label{sec:cIndex}

The area under the ROC curve (AUC) is a widely used summary measure of predictive model performance in many disciplines, including medicine. However, for the same reasons that standard classification techniques fail on censored outcomes, the AUC can also be misleading. Hence, we propose to employ a generalization of the AUC, the concordance index (C-index) for censored data. In the absence of censoring, the C-index and AUC coincide; however, the C-index is more easily adapted to account for censoring.

As described in \cite{Harrell}, the C-index adapted for censoring considers the concordance of survival outcomes versus predicted survival probability among pairs of subjects whose survival outcomes can be ordered; namely, among pairs where both subjects are observed to experience a CV event, or one subject is observed to experience a CV event  before the other subject is censored. Pairs in which both subjects are censored or in which the censoring time of one precedes the failure of the other do not contribute to this metric. Let $\hat{P}_i(E|{\mbf{X}})$ be the estimated probability that the $i^{th}$ subject experiences an event within $\tau$ years. The the C-index is given by
\begin{equation}
C_{cens}(\tau) = \frac{ \sum_{i \neq j}  \delta_i \indicatorBig{V_i<V_j}\indicatorBig{\hat{P}_i(E|{\mbf{X}})<\hat{P}_j(E|{\mbf{X}})}}{ \sum_{i \neq j} \delta_i \indicatorBig{V_i<V_j} }
\end{equation}

Note that the only pairs which contribute to $C_{cens}(\tau)$ are those where one subject experiences an event prior to $\tau$ and the other is known not to have experienced an event before the first subject.

\subsubsection{Net Reclassification Improvement}\label{sec:NRI}

The C-index may be inadequate to distinguish between models that differ in relatively modest but clinically important ways \citep{pencina2008evaluating}. Therefore, in addition to the C-index, we also evaluate the performance of our models using the Net Reclassification Improvement (NRI) metric  \citep{pencina2008evaluating}, which allows the incorporation of relevant domain knowledge into the performance evaluation process.  As with the other metrics presented in this section, the NRI must be adapted to handle censored outcomes.

The NRI compares the number of ``wins'' for two models among discordant predictions. It has been argued that NRI is a particularly relevant measure of comparison between models in the clinical domain, where it is often more important to discriminate between lower and higher risk patients than to estimate their risk precisely. Briefly, the NRI is computed by cross-tabulating predictions from two different models with table cells defined by clinically meaningful cardiovascular risk categories or bins, then comparing the agreement of discordant predictions with actual event status. Formally, the NRI for comparing prediction models $M_1$ and $M_2$ using fully observed (i.e., not censored) binary event data is given by:
\begin{equation}\label{eq:NRI}
\mathrm{NRI}(M_1,M_2) = \frac{E_{M_1}^{\uparrow} - E_{M_2}^{\uparrow}}{n_E} + \frac{ \bar{E}_{M_1}^{\downarrow} - \bar{E}_{M_2}^{\downarrow}}{ n_{\bar{E}}}
\end{equation}

Here $E_{M_1}^{\uparrow}$ is the number of individuals who experienced events and were placed in a higher risk category by $M_1$ than $M_2$ (i.e., a number of ``wins'' for $M_1$ over $M_2$ among patients who had events), and the opposite change in risk categorization yields $E_{M_2}^{\uparrow}$). Similarly, $\bar{E}_{M_1}^{\downarrow}$ and $\bar{E}_{M_2}^{\downarrow}$ count the number of individuals who did not experience an event and were ``down-classified'' by $M_1$ and $M_2$, respectively (i.e., ``wins'' among patients who did not have events). Also, $n_E$ and $n_{\bar E}$ are the total number of patients with events and non-events, respectively.  A positive $\mathrm{NRI}(M_1,M_2)$ means better reclassification performance for $M_1$, while a negative $\mathrm{NRI}(M_1,M_2)$ favors $M_2$. While NRIs can theoretically range from -1 to 1, in the risk prediction setting they do not typically exceed the range [-0.25, 0.25].  For example, \cite{Cook2009} calculated the effects of omitting various risk factors from the Reynolds Risk Score model for prediction of 10-year cardiovascular risk. The estimated NRIs ranged from -0.195 (omitting age) to -0.032 (omitting total cholesterol or parental history of MI), and all were statistically significant at the 0.05 level. In our application, we defined three risk strata based on clinically relevant cutoffs for the risk of experiencing a cardiovascular event within 5  years: 0-5\% (low risk), 5-10\% (moderate risk), and $>$ 10\% (high risk); risk predictions for an individual were considered discordant between two models if the predictions fell in different ranges.

The predictions that are reclassified from one risk category to another can be represented in two tables, one for people having events (Table~\ref{table:NRItabEvt}) and one without (Table~\ref{table:NRItabNoevt}), where the risk has been categorized into 3 levels: ``high'', ``medium'', and ``low''. The high, medium, and low categories for CV risk are usually defined as risk intervals, such as 10-100\%, 5-10\%, and 0-5\%. The entry in Table~\ref{table:NRItabNoevt}, $Nn_{ij}$, means the number of people without events who were classified as belonging to risk category $i$ by model $M_2$ that were reclassified as belonging to risk category $j$ by model $M_1$.  Similarly, the entry in Table~\ref{table:NRItabEvt}, $Ne_{ij}$, means the number of people with events who were classified as belonging to risk category $i$ by model $M_2$ that were reclassified as belonging to risk category $j$ by model $M_1$.

For a person with events, a ``win'' for model $M_1$ over $M_2$ means $M_1$ predicts a higher risk category for the person than $M_2$.  Based on our reclassification table,  $E_{M_1}^{\uparrow} = Ne_{mh}+Ne_{lh}+Ne_{lm}$. Similarly, the number of ``wins'' $M_2$ has over $M_1$ for people with events is: $E_{M_2}^{\uparrow} = Ne_{hm}+Ne_{hl}+Ne_{ml}$. For people without events, we can write the write the corresponding ``wins'' from Table~\ref{table:NRItabNoevt} as follows:  $\bar{E}_{M_1}^{\downarrow} = Nn_{hm}+Nn_{hl}+Nn_{ml}$ and $\bar{E}_{M_2}^{\downarrow} = Nn_{mh}+Nn_{lh}+Nn_{lm}$. $n_E$ and $n_{\bar{E}}$ are the sum of the entries in Tables~\ref{table:NRItabEvt} and~\ref{table:NRItabNoevt} respectively. From the values evaluated using the reclassification tables, the Net Reclassification improvement is evaluated using Equation~\eqref{eq:NRI}.

\begin{table}
\caption{Net Reclassification Improvement computation tables for comparing  performance of model $M_1$ vs. model $M_2$.}
\begin{subtable}[b]{1.0\textwidth}
\centering
\caption{Reclassification table for people who do not have CV events.}
\begin{tabular} {|l|l||l|l|l|}\hline
\multicolumn{2}{|c||}{}  &\multicolumn{3}{|c|}{Model $M_1$} \\ \cline{2-5}
			& Risk              &  high($h$)  & medium($m$)     &  low($l$)       \\ \hline\hline
			& high($h$)         &  $Nn_{hh}$  &  $Nn_{hm}$      &    $Nn_{hl}$    \\ \cline{2-5}
Model $M_2$ & medium($m$)       &  $Nn_{mh}$  &  $Nn_{mm}$      &    $Nn_{ml}$    \\ \cline{2-5}
			& low($l$)          &  $Nn_{lh}$  &  $Nn_{lm}$      &    $Nn_{ll}$    \\ \hline
\multicolumn{5}{c}{}\\
\end{tabular}
\label{table:NRItabNoevt}
\end{subtable}
\begin{subtable}[b]{1.0\textwidth}
\centering
\caption{Reclassification table for people who have CV events.}
\begin{tabular} {|l|l||l|l|l|}\hline
\multicolumn{2}{|c||}{}  &\multicolumn{3}{|c|}{Model $M_1$} \\ \cline{2-5}
			& Risk              &  high($h$)   & medium($m$)     &  low($l$)      \\ \hline\hline
			& high($h$)         &  $Ne_{hh}$   &    $Ne_{hm}$    &  $Ne_{hl}$     \\ \cline{2-5}
Model $M_2$ & medium($m$)       &  $Ne_{mh}$   &    $Ne_{mm}$    &  $Ne_{ml}$     \\ \cline{2-5}
			& low($l$)          &  $Ne_{lh}$   &    $Ne_{lm}$    &  $Ne_{ll}$     \\ \hline
\end{tabular}
\label{table:NRItabEvt}
\end{subtable}
\end{table}

The NRI statistic cannot be applied directly to data where the outcome status of some subjects is unknown. In our setting, omitting subjects with less than five years of follow-up (or treating them as non-events) will result in biased estimates of the NRI.  To evaluate risk reclassification on our test data which are subject to censoring, we use a ``censoring-adjusted'' NRI (cNRI) due to \cite{Pencina_2011} which takes the form:

\begin{equation}
\mathrm{cNRI}(M_1,M_2) = \frac{E_{M_1}^{*,\uparrow} - E_{M_2}^{*,\uparrow}}{n^*_E} + \frac{ \bar{E}_{M_1}^{*,\downarrow} - \bar{E}_{M_2}^{*,\downarrow}}{ n^*_{\bar{E}}}
\label{eq:cNRI}
\end{equation}

\noindent where $E_{M_1}^{*,\uparrow}, E_{M_1}^{*,\downarrow}, E_{M_2}^{*,\uparrow}, E_{M_2}^{*,\downarrow}, n^*_E$ and $n^*_{\bar E}$ are analogous to the quantities in \eqref{eq:NRI}, but correspond to expected number of subjects in each category, with the expectations computed using the Kaplan-Meier estimator to account for censoring. As with the usual NRI, a positive $\mathrm{cNRI}(M_1,M_2)$ means better reclassification performance for model $M_1$, while a negative $\mathrm{cNRI}(M_1,M_2)$ favors model $M_2$. Uncertainty intervals can be obtained by computing the cNRIs on the test set for predictions fitted to 500 bootstrap resamples of the training data.

\section{Electronic health data and preprocessing}\label{sec:data}

Our study was conducted utilizing the HMO Research Network Virtual Data Warehouse (HMORN VDW) from a healthcare system from the Midwestern U.S. The VDW stores data in standardized data structures including insurance enrollment, demographics, pharmaceutical dispensing, utilization, vital signs, laboratory, census and death records. These data are obtained from both the EMR and insurance claims. This health care system includes both an insurance plan and a medical care network in an open system which is partially overlapping. That is, patients of the insurance plan may be served by either the internal medical care network and or by external healthcare providers, and the medical care network serves patients within and outside of the insurance plan.  Patient-members who do not visit any of the clinics and hospitals in-network do not have any medical information (e.g., blood pressure information) included in the EMR of this system. Furthermore, once the patient-member disenrolls from the HMO, the patient no longer has any information recorded in the EMR or insurance claims data.

Study population was initially selected from those enrolled in the insurance plan between 1999 and 2011 and who had at least one outpatient medical encounter at an ``in-network'' clinic. This initial selection identified 448,306 subjects.

The goal of our analysis is to develop accurate prediction models of CV risk for patients seen in the primary clinics of the HMO.  Such models will help inform physicians and patients of appropriate courses of action to take in order to reduce the patient's likelihood of experiencing a CV event.

In many epidemiological cohorts, patients are screened at a single or small number of visits, where measurements on all risk factors are collected, and then followed from the final screening visit to determine if and when they experienced a CV event.  However, in clinical practice, patients may have relevant risk factors recorded over a series of visits to the physician.  Therefore, we divided the EHD on any given patient-member into: (i) a baseline period, where we ascertained the risk factors, and (ii) a follow-up period, where we assessed whether a patient experienced a CV event (and, if so, when).  The baseline period consisted of the time between the first blood pressure reading during the enrollment period and the date of the final blood pressure reading at most 1.5 years from the first measurement.  This approach balances the competing goals of having a long baseline period so that we capture data on as many features as possible and a long follow-up period to reduce censoring.

The follow-up period for a patient begins at the end of the baseline period and continues until either the patient experiences a CV event (defined below), or the patient disenrolls from the HMO for more than 90 days, or the study ends (2011), whichever comes first.

\subsection{Inclusion and exclusion criteria}

To ensure that we had sufficient time to collect baseline risk factors on subjects, we restricted the analysis to those subjects with at least one year of continuous insurance enrollment. Some of the patients were sporadically enrolled during the period of study; however, for the purpose of our analysis, we ignored gaps in enrollment less than 90 days and considered a patient-member continuously enrolled over this period (these gaps in enrollment are likely due to administrative errors or patients changing employers but still electing coverage with the same HMO).

We further restrict our study population to include only patients with two medical encounters in the in-network clinic with blood pressure information at least 30 days but at most 1.5 years apart, with drug coverage, and greater than 18 years of age at the end of the baseline period.  These inclusion criteria were implemented because we wanted to predict CV risk among those patients treated routinely in the primary care clinic. Patients who are only infrequently treated in the emergency room or urgent care clinics (i.e., settings where patients are unlikely to be counselled about their CV risk) were not of interest in this analysis.

\begin{figure}[h]
\centering
\includegraphics[width=0.50\textwidth]{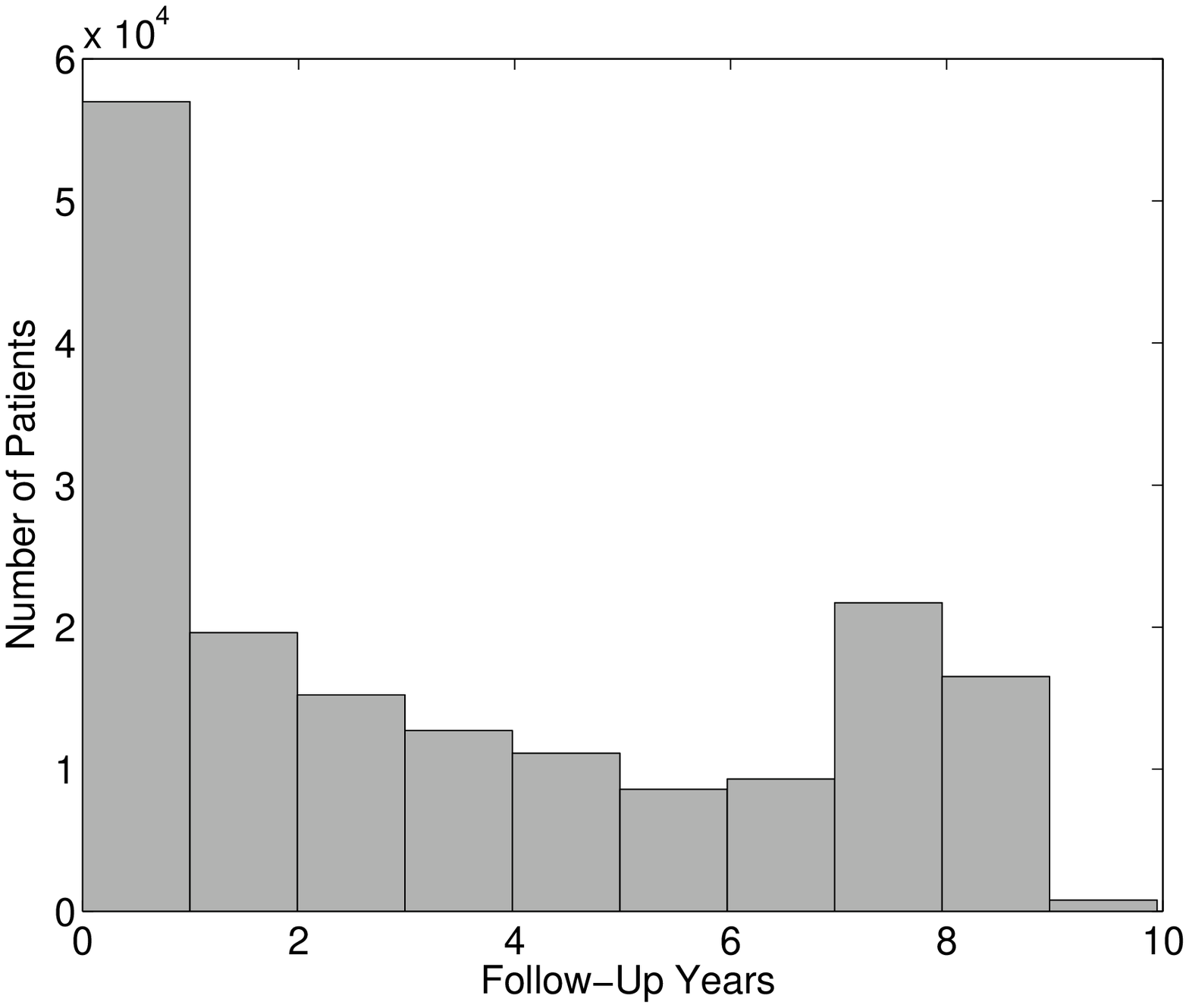}
\caption{Distribution of patient follow-up times, i.e., time from the end of the baseline period until the patient experiences a CV event, the patient disenrolls from the HMO for more than 90 days, or the study ends, in our entire cohort after applying inclusion and exclusion criteria.}
\label{figure:followDist}
\end{figure}

Finally, in this study we included only non-diabetic patients.  Diabetic patients represent a highly specialized population and would benefit from a specialized risk prediction model that is targeted specifically to them (such as the UKPDS model \citep{clarke2004model}), which was beyond the scope for this specific paper. The inclusion and exclusion criteria are summarized in Figure~\ref{figure:CONSORT}, and the distribution of the follow-up periods for the resulting analysis cohort is given in Figure~\ref{figure:followDist}.

\begin{center}
  \sffamily
  \footnotesize
  \begin{tikzpicture}[auto,
    block_center/.style ={rectangle, draw=black, thick, fill=white,
      text width=16em, text centered,
      minimum height=4em},
    block_left/.style ={rectangle, draw=black, thick, fill=white,
      text width=16em, text ragged, minimum height=4em, inner sep=6pt},
    block_noborder/.style ={rectangle, draw=none, thick, fill=none,
      text width=18em, text centered, minimum height=1em},
    block_assign/.style ={rectangle, draw=black, thick, fill=white,
      text width=18em, text ragged, minimum height=3em, inner sep=6pt},
    block_lost/.style ={rectangle, draw=black, thick, fill=white,
      text width=16em, text ragged, minimum height=3em, inner sep=6pt},
      line/.style ={draw, thick, -latex', shorten >=0pt}]
    \matrix [column sep=5mm,row sep=3mm] {
      \node [block_center] (initial) {Initial cohort (n=448,306).};
      & \node [block_left] (included1) {Included patients-members: \\
        a) Enrolled between 1999-2011 AND \\
        b) At least one visit to in-network clinic or hospital}; \\
      \node [block_center] (exclude1) {Exclude 87,557 patients with continuous enrollment less than one year.};
      & \\
       \node [block_center] (exclude2) {Exclude 160,977 patients without two blood pressure measurements at least 30 days and at most 1.5 years apart.};
      & \\
     \node [block_center] (exclude3) {Exclude 4,743 patients without drug coverage.};
      & \\
         \node [block_center] (exclude4) {Exclude 6,269 patients under 18 years of age.};
      & \\
         \node [block_center] (exclude5) {Exclude 16,186 diabetic patients.};
      & \\
         \node [block_center] (final) {Final analysis cohort (n=172,571).};
      & \\ };
    \begin{scope}[every path/.style=line]
      \path (initial)   -- (included1);
      \path (initial)   -- (exclude1);
      \path (exclude1) -- (exclude2);
      \path (exclude2) -- (exclude3);
      \path (exclude3) -- (exclude4);
      \path (exclude4) -- (exclude5);
      \path (exclude5) -- (final);
    \end{scope}
  \end{tikzpicture}
  \captionof{figure}{Flowchart of inclusion and exclusion criteria for analysis }
  \label{figure:CONSORT}
\end{center}

\subsection{Risk factor ascertainment}

Risk factors incorporated in the Bayesian network include age, gender, systolic blood pressure (SBP), smoking status, body mass index (BMI), cholesterol-related measurement values (LDL, HDL, TRG), blood pressure and cholesterol medications, as well as indicators of pre-existing cardiovascular disease (i.e., pre-existing related diagnoses or procedures).  The summary statistics for the risk factors are given in Table \ref{table:cohortProps}. We also provide a brief discussion on how each of the features and the outcome, CV event, was defined based on information in the EMR and claims data.

\begin{center}
\begin{sidewaystable}
\caption{Summary measures of the risk factors included in our prediction models in the entire study cohort.}
\begin{tabular}{l|ccl}\hline \noalign{\smallskip}
                          & {\bf{Median (IQR)}}       &                   &                                                                          \\
{\bf{Feature Name}}       &      or                   & {\bf{\% Missing}} &  {\bf{Description}}                                                      \\
                          & {\bf{N (\%)}}             &                   &                                                                          \\ \noalign{\smallskip}\hline\noalign{\smallskip} \hline
{\bf{Gender}}             &                           &                   &                                                                          \\
\ \ \ Female              & 105,234 (60.98)           &  0                &                                                                          \\
\ \ \ Male                &  67,337 (39.02)           &  0                &                                                                          \\
{\bf{Age }}(Years)        & 42.00   (30.00  -  55.00) &  0                &  Age  at the end of the baseline period    \\
{\bf{SBP }}(mm Hg)        & 118.00  (110.00 - 128.00) &  0                &  Mean systolic blood pressure  during baseline period    \\
{\bf{BMI }}(kg/m$^2$)     & 26.72   (23.60  -  31.06) &  10               & Body mass index                                                \\
{\bf{LDL }}(mg/dL)        & 118.00  (96.00  - 142.00) &  66               &  Final low density  lipoprotein during baseline period   \\
{\bf{HDL }}(mg/dL)        & 48.00   (39.00  -  58.00) &  55               &  Final high density  lipoprotein during baseline period  \\
{\bf{TRG }}(mg/dL)        & 108.00  (77.00  - 158.00) &  66               &  Final tryglyceride  during baseline period              \\
{\bf{Smoking}}            &                           &                   & Smoking status in EMR                                           \\
\ \ \ Never or Passive    & 126,463 (73.28)           &   0               &                                                                          \\
\ \ \ Quit                & 16,898  (9.79)            &   0               &                                                                          \\
\ \ \ Current             & 29,210  (16.93)           &   0               &                                                                          \\
{\bf{Comorbidity}}        &                           &                   & Presence of comborbidities related to cardiovascular disease  \\
\ \ \ Yes                 &  14,547 (8.43)            &   0               &                                                                          \\
\ \ \ No                  & 158,024 (91.57)           &   0               &                                                                          \\
{\bf{SBP Meds}}           &                           &   0               &  Number of SBP medication classes filled during baseline period          \\
\ \ \ 0                   & 118,685 (68.77)           &   0               &                                                                          \\
\ \ \ 1                   & 20,613 (11.94)            &   0               &                                                                          \\
\ \ \ 2                   & 13,883 ( 8.04)            &   0               &                                                                          \\
\ \ \ 3+                  & 19,390 (11.24)            &   0               &                                                                          \\
{\bf{LDL Meds}}           &                           &   0               &            \\
\ \ \ 0                   & 154,972 (89.80)           &   0               &   Number of LDL medication classes filled during baseline period                                                                       \\
\ \ \ 1+                  & 17,599 (10.20)            &   0               &                                                                          \\ \noalign{\smallskip}
\hline
\end{tabular}
\label{table:cohortProps}
\end{sidewaystable}
\end{center}

{\textbf{Systolic blood pressure (SBP):}} Calculated as an average of all the blood pressure measurements taken during the baseline period. Blood pressure readings obtained during emergency department visits, urgent care visits, hospital admission, and during procedures (e.g., surgeries) were excluded from consideration because they may be influenced by acute conditions.

{\textbf{Body mass index (BMI):}} Calculated as a function of patient's height and weight. The height of an individual is the average height measured at any encounter (possibly outside of the baseline period). Because all subjects in the analysis dataset are over 18 years of age, we expect height to remain relatively constant over the follow-up period.  The weight is calculated as an average of all weight measurements taken during the baseline period.

{\textbf{Low density lipoprotein (LDL), high density lipoprotein (HDL), and triglycerides (TRG):}} The most recent laboratory measurements before the end of the baseline period is used for these lipid measures including low density lipoprotein, high density lipoprotein, and triglycerides.

{\textbf{Smoking status:}} Information about smoking is complicated by the fact that many individual's responses vary considerably over time.  In our dataset, there are four categories of smoking history, never smoked, smoking, quit smoking, and passive (i.e., second-hand) smoking.  In our analysis, a person is considered to have never smoked only if they consistently recorded ``no smoking'' throughout their association with the insurance provider. A person who has recorded at least two ``smoking'' responses is considered currently smoking.  For the purpose of constructing the model we combine the ``passive smoking'' and ``no smoking'' categories.

{\textbf{SBP and LDL medications:}} In our model, SBP medications are represented as the number of different medication categories a person is prescribed at the end of the baseline period. In particular, SBP medication categories included: alpha-blockers, beta-blockers, calcium-blockers, ace-inhibitors, angiotensin, vasodialator, and diuretics. LDL medications represents an indicator for whether or not a patient is taking any LDL lowering medications, such as statins and fibrates, at the end of the baseline period.  For our analysis, we ignore information regarding the specific drug dosages because it is difficult to make comparisons between doses from different variants of the same drug.


{\textbf{Comorbidities:}} Comorbidities represent serious pre-existing conditions (diseases) and previously occurred CV events or procedures (surgeries).  The existence of comorbidities significantly elevate the risk of having a CV event in the future.  In our study, we included the presence of any of the following diagnoses (including a diagnosis of a ``history of'' these conditions)  at any point before the end of the baseline period: chronic kidney disease, coronary heart disease, cardiovascular disease, peripheral artery disease, atrial fibrillation, congestive heart failure, MI, and stroke. As we discuss below, the diagnosis may be part of the EMR or contained as part of an insurance claim. The Bayesian model that we consider treats comorbidities as a binary variable.

{\textbf{CV Event (outcome variable):}}   Events are the first recorded stroke, myocardial infarction (MI), or procedure proximal to stroke or MI (e.g., coronary artery bypass surgery, stent for either the coronary arteries or carotid artery) after the baseline period. This information is obtained from diagnosis codes recorded by physicians or inferred from procedures such as bypass surgery or stent placement performed on an individual. In addition to using procedure and diagnosis codes to infer if a CV event occurred, we consider a patient to have experienced a CV event if the cause of death listed on the death certificate included MI or stroke.

We note that the diagnosis and procedure codes used to define CV events and comorbidities may be part of the EMR or part of claims data (to justify the insurance claim). The implication of this is that patients do not have to seek care at an in-network hospital following a CV event to infer that the patient had a CV event.  The total number of first CV events recorded within the follow-up period is 4,504; the Kaplan-Meier 5-year event rate for the entire analysis cohort is 2.61\%.


\subsection{Determining the structure of the Bayesian network}

Figure \ref{figure:BayesNet} displays the structure of the Bayesian network that we used to construct our prediction models. The structure was determined by combining known relationships from the medical literature with input from our clinical colleagues. For example, SBP is known to depend on age, BMI, and the number of blood pressure medication classes prescribed~\citep{rockwood2011blood}; therefore, age, BMI, and blood pressure medication nodes are represented as parents of SBP nodes in the DAG. Also, LDL, HDL, triglycerides (TRG), age, and LDL medications are similarly known to interact \citep{grover2003improving}.  In this case, the nodes corresponding to age and LDL medications are the parents of the lipid nodes, which are modelled jointly.

In some cases, it may be preferable to consider several network structures reflecting plausible relationships between predictor variables. In the process of developing our approach, we considered several candidate network structures; within a set of similar candidates, the particular structure did not have a large impact on prediction performance.

\section{Models and evaluation metrics} \label{sect:data-analysis}

Our goal is to build risk models from EHD and, in the process, address issues typically encountered with EHD, such as right-censored outcomes, non-linear and non-monotonic effects of risk factors on risk of events, and overfitting in subgroups that are represented by relatively few patients.  In the context of estimating the five-year risk of a cardiovascular event using the data described in Section~\ref{sec:data}, we sought to compare our approach  which trains the Bayesian network in Figure~\ref{figure:BayesNet} using inverse probability of censoring weights (Section~\ref{sect:IPCW}) and using model averaging (Section~\ref{sect:MA}) to other more traditional Bayesian network approaches as well as other techniques for modeling censored outcomes. Our approach, which we refer to as Bayes-AC as it includes both model {\bf{a}}veraging and {\bf{c}}ensoring weights,  allows for non-linear relationships between the risk factors and outcome, addresses censored outcomes, and protects against overfitting in small subpopulations. In addition to our main approach, we also considered a more basic Bayesian model using a single multivariate normal distribution (Bayes-1C) to model $P(\mbf{G}_i|E)$, i.e., no model averaging, but still accounting for censored outcomes.

For comparison, we train the Bayesian network using an {\em{ad hoc}} approach for censored observations in which we excluded patients who did not experience an event and did not have 5-years of follow-up. This {\em{ad hoc}} approach to censoring was used both when the number of mixture components was fixed (Bayes-1 and Bayes-3) and with model averaging (Bayes-A).

In addition to the different variations of the Bayesian network models, we considered a Cox proportional hazards model (COX) because this model is well-known in the medical community and well-suited to work with censored data.   A drawback of the proportional hazards model is that it does not automatically allow for non-linear relationships between the risk factors and log hazard. We chose to parameterize the proportional hazards model using the same approach as \cite{DAgostino_2008}. One of the limitations in using the Cox proportional hazards model is that it requires complete data to fit the model and to predict risks, therefore, we have to impute the missing data.  We make use of a k-nearest neighbor based algorithm to impute missing data \citep{crookston2008yaimpute}. The data is conditioned on gender and comorbidity status before it is imputed. This imputation approach requires at least one of the attributes for a given data point to be known. In our case, age and SBP are known for all patients; therefore, this method of imputation can be easily applied.

Brief descriptions of the models that we have used in our analysis are given in Table~\ref{table:modelDisc}.  All models are trained on 129,428 patients (75\% of the entire analysis population) drawn at random from the cohort. The performance of the model is evaluated based on the risk predictions of the remaining 43,143 patients. The models are compared based on their calibration (as described in Section \ref{sec:calibration}) and discrimination metrics, i.e., the C-index and cNRI (as described in Sections \ref{sec:cIndex} and \ref{sec:NRI}). In addition to the calibration and the discrimination metrics, we also plot the average risk predicted by the model against the observed risk across different patient groups. While the calibration statistic K helps us compare the models using a single numeric metric, these plots provide us with more information regarding the calibration of a model for different risk ranges and the direction of deviation (over- or under-prediction) of the model from the observation. A perfectly calibrated model would be indicated by a 45 degree line in these plots.

To construct these plots, we bin patients based on their predicted risk and evaluate the Kaplan-Meier estimate of the 5-year probability of experiencing a CV event.  For our plots, the patients in the test set are partitioned into 5 risk bins with boundaries determined by four equally spaced points between the 5th and 95th percentile of predicted risk in the test set.  Note that the predicted and the observed risk of the patients in the first and the last bin include people whose predicted risk is lesser than the 5 percentile and the 95 percentile risk predictions respectively. We also evaluate the $95\%$ confidence interval of the Kaplan-Meier estimate of the observed risk, represented as error bars in the plots.

\begin{table}[ht]
\centering
{\small
\caption{ Overview of the models considered in the analysis of 5-year cardiovascular event risk.}
\begin{tabular}{ll} \hline\noalign{\smallskip}
\hspace{1em}{\bf{Model Name}}  & {\bf{Description}}  \\ \noalign{\smallskip}\hline\noalign{\smallskip}
\hline
\multicolumn{2}{l}{Bayesian network models without accounting for censoring} \\
\hspace{1em}{\bf{Bayes-1}}     & basic model, using 1 normal distribution for all continuous variables \\
\hspace{1em}{\bf{Bayes-3}}     & more complex model, using mixture of 3 normal distributions for all continuous variables   \\
\hspace{1em}{\bf{Bayes-A}}     & ensemble approach, using model averaging for models with mixtures of different sizes (1-4 normals)    \\
\multicolumn{2}{l}{Bayesian network models accounting for censoring} \\
\hspace{1em}{\bf{Bayes-1C}}    & Bayes-1 with inverse probability of censoring weights                      \\
\hspace{1em}{\bf{Bayes-AC}}    & Bayes-A with inverse probability of censoring weights  \\
\multicolumn{2}{l}{Standard regression-based baseline approach for censored data}  \\
\hspace{1em}{\bf{COX}}         & Cox proportional hazards model                                        \\ \noalign{\smallskip}\hline
\end{tabular}
\label{table:modelDisc}
}
\end{table}

\section{Results} \label{sect:data-analysis2}

The calibration and discrimination of the models, evaluated on the test set, is summarized in Table~\ref{table:resSummary}. Specific comparisons between our approach, which incorporated model averaging and IPCW to train a Bayesian network, and other more standard modeling approaches are described in detail below.

\begin{table}[ht]
\centering
\caption{Calibration and discrimination of the models described in Table~\ref{table:modelDisc}, evaluated on the hold-out test set. {\em{Predicted event rate}}: Average predicted probability of experiencing a CV event within 5 years, the Kaplan-Meier estimate in the test set was 2.61\%;  {\em{Calibration}}: calibration test statistic K; {\em{C-index}}: Concordance index; {\em{cNRI}}: net reclassification improvement for censored outcomes.   }
\begin{tabular}{lcccc} \hline\noalign{\smallskip}
                    &  Predicted event rate (\%) & Calibration&   C-index  & cNRI (\%)         \\
                    & (Observed rate: 2.61\%)    &  statistic K&            &(compared to COX)  \\ \noalign{\smallskip}\hline\noalign{\smallskip}
Bayes-1             &   5.60                     &  0.223     &   0.837    &   10.86           \\
Bayes-3             &   5.34                     &  0.288     &   0.821    &    6.76           \\
Bayes-A             &   5.34                     &  0.186     &   0.843    &   10.37           \\
Bayes-1C            &   2.67                     &  0.011     &   0.843    &    8.18           \\
Bayes-AC            &   2.45                     &  0.012     &   0.849    &    8.88           \\
COX                 &   2.12                     &  0.017     &   0.839    &     -             \\ \noalign{\smallskip}\hline
\end{tabular}
\label{table:resSummary}
\end{table}

\subsection{Censoring-unaware Bayesian networks}
\begin{figure}[ht]
\centering
\includegraphics[width=0.45\textwidth]{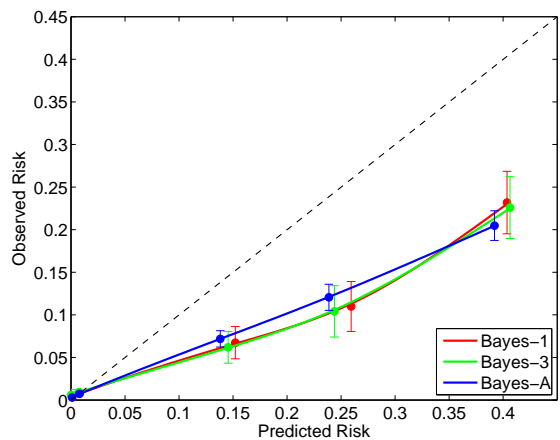}
\caption{Calibration of Bayesian network models without IPCW both with and without model averaging on the hold-out test set.}
\label{figure:CalibNoCens}
\end{figure}

The Bayesian network models that do not incorporate inverse probability of censoring weighting discard subjects who do not have a follow-up time of at least 5 years and do not experience an event, but still include subjects who have events even if they have a follow-up time of less than five years. As a result, all of these models (Bayes-1, Bayes-3, and Bayes-A) over-predict risk and hence are poorly calibrated (see Figure~\ref{figure:CalibNoCens}).  This estimation approach leads to a poorly calibrated model which over-predicts risk significantly. Although the observed 5 year event rate for the test set was 2.61\%, the average risk predicted by Bayes-1, Bayes-3, and Bayes-A was 5.60\%, 5.34\%, and 5.34\%, respectively.

The Bayes-3 model, in which all continuous features are modeled as a mixture of three multivariate normal distributions, consistently performs worse than the modeling approach with less flexibility (Bayes-1) across all measures of calibration and discrimination. In contrast, considering a model-averaged estimate of CV risk, which averages over model complexity using a data-driven approach, led to improvements in both calibration and discrimination. Specifically, the averaged model predicts the risk of a CV event better in groups where few events occur. In the sub-population of patients that are not on blood pressure medications and has an event rate of 0.39\% (as compared to an average event rate of 2.61\%), the Bayes-A model has C-index of 0.75 which is significantly higher than the C-indices 0.65 and 0.56 for the Bayes-1 and Bayes-3 models, respectively. In summary, complex model strategies are able to extract more structure from the data but must be implemented intelligently. The benefit of model averaging is that it allows the analyst to consider more complex models but these more complex models (with greater number of mixture components) are only given substantial weight in Equation~\ref{eq:model_avg_weight} if there is enough improvement in the model fit to justify models with more parameters.  As mentioned earlier, we use the BIC as the metric to balance model complexity and parsimony.

We note briefly that, while these models show a significant improvement in the discrimination compared to the standard Cox proportional hazards models using cNRI, it is well known that the NRI statistic can be misleading when one of the models is badly calibrated, such as the models that do not account for censoring \citep{Pepe_2011}.  The cNRI statistic that we consider weights the percentage of ``wins'' for those experiencing an event equally to those not experiencing an event. The Bayesian network models that do not account for censoring are heavily biased toward predicting higher risk; therefore, it produces a significant number of ``wins'' for people who have events. Therefore, it is most meaningful to use the cNRI metric only when both models being compared are reasonably well-calibrated.

\subsection{Bayesian networks accounting for censoring using IPCW}
\begin{figure}[ht]
\centering
\includegraphics[width=0.45\textwidth]{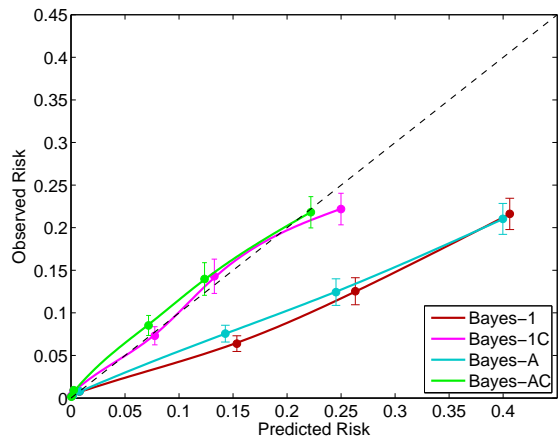}
\caption{Calibration of censoring-aware vs. censoring-unaware Bayesian network models on the hold-out test set.}
\label{figure:CalibNoCensBayes}
\end{figure}

Comparing the parsimonious Bayesian networks and model-averaged Bayesian networks that do and do not incorporate inverse probability of censoring weights (Bayes-1 versus Bayes-1C and Bayes-A versus Bayes-AC), there is a dramatic improvement in the calibration of those models which properly account for censoring (Figure \ref{figure:CalibNoCensBayes}). In these models, the average predicted event rates 2.67\% (Bayes-1C) and 2.45\% (Bayes-AC) is much closer to the observed event rate in the test set than the equivalent models that are fit ignoring censoring. The calibration statistic is also much closer to zero, the value we would expect for a perfectly calibrated model. Both Bayes-1C and Bayes-AC also show slightly improved discrimination compared to the respective models that do not account for censoring (C-index 0.837 versus 0.843 and 0.843 versus 0.849, respectively). Considering the cNRI, both of these models are significantly better in terms of discrimination than the standard Cox proportional hazards model but, unlike the Bayes-1 and Bayes-A models, do not sacrifice calibration to improve net reclassification.

Overall, the Bayes-AC model, which uses inverse probability of censoring weights and model averaging, results in the best performance of all our Bayesian network models.  Unlike the Bayes-1 model (a more traditional Bayesian network model), our approach properly accounts for censoring leading to improved calibration and properly protects against over-fitting leading to improved discrimination.

\subsection{Censoring-aware Bayesian networks vs. traditional survival analysis}
As noted previously, the Cox proportional hazards model (COX) is the standard approach for data in which the outcome may be right-censored.  As compared to this model, our Bayesian network model (Bayes-AC) provides improvement across all calibration and discrimination metrics considered for this predictive task. Overall, the average predicted event rate for the Bayes-AC model (2.48\%) was closer to the observed event rate (2.61\%) in the test set than the one demonstrated by COX (2.12\%).  The C-index of Bayes-AC is 0.849 (versus 0.839 for COX), and the cNRI of Bayes-AC compared to COX is
8.88\%, both of which reflect significant improvement in the discrimination.  To put the reclassification performance in context, this is more than double the improvement that can be obtained from adding the total cholesterol (or about half the improvement from adding age) into the COX risk prediction model \citep{Cook2009}.  Our improvement in cNRI can largely be attributed to the fact that Bayes-AC predicts a higher risk category for substantially more people with events compared to COX (Tables~\ref{table:NRItabEvt2} and~\ref{table:NRISummary}).   On the other hand, the COX model predicts a lower risk category for a similar percentage of patients without events as Bayes-AC (Tables~\ref{table:NRItabNoevt2} and~\ref{table:NRISummary}).  More generally, the COX model tends to under-predict the risk of events as compared to Bayes-AC largely due to its comparatively poorer calibration.  Overall, Bayes-AC reclassifies a higher fraction of people in the correct direction.

\begin{figure}[ht]
\centering
\includegraphics[width=0.45\textwidth]{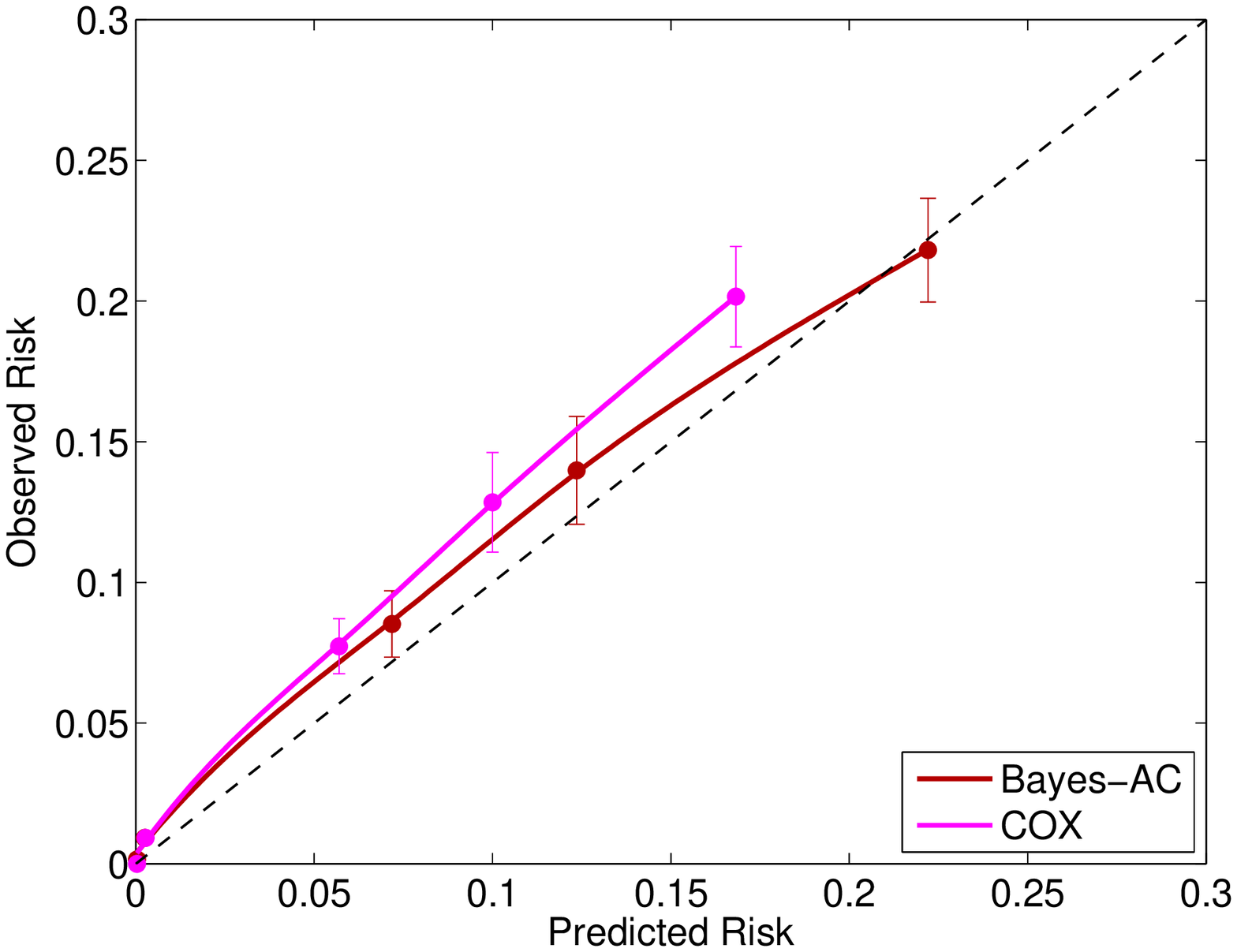}
\caption{Calibration comparison of the Cox proportional hazards model to the censoring-aware Bayesian network with model averaging on the hold-out test set.}
\label{CalibCensBayesCox}
\end{figure}

\begin{table}
\caption{Censoring-adjusted Net Reclassification Improvement computation tables for comparing performance of Bayes-AC vs. COX on the hold-out test set.}
\begin{subtable}[b]{1.0\textwidth}
\centering
\caption{Reclassification table for patients without CV events.}
\begin{tabular} {|l|l||c|c|c|}\hline
\multicolumn{2}{|c||}{}  &\multicolumn{3}{|c|}{Bayes-AC} \\ \cline{2-5}
			& Risk              &  $>$10\%     &  5-10\%     &  0-5\%       \\ \hline\hline
			& $>$10\%            &  2,091       &   490           &    240          \\ \cline{2-5}\cline{2-5}
   COX      & 5-10\%        &   878       &   808           &    1,023         \\ \cline{2-5}
			& 0-5\%          &   307       &   729           &    31,058        \\ \hline
\multicolumn{5}{c}{}\\
\end{tabular}
\label{table:NRItabNoevt2}
\end{subtable}
\begin{subtable}[b]{1.0\textwidth}
\centering
\caption{Reclassification table for patients with CV events.}
\begin{tabular} {|l|l||c|c|c|}\hline
\multicolumn{2}{|c||}{}  &\multicolumn{3}{|c|}{Bayes-AC} \\ \cline{2-5}
			& Risk             &  $>$10\%       &  5-10\%     &  0-5\%      \\ \hline\hline
			& $>$10\%           &  513         &    52           &  14            \\ \cline{2-5}\cline{2-5}
   COX      & 5-10\%       &  121         &    61           &  36            \\ \cline{2-5}
			& 0-5\%         &  34          &    47           &  196           \\ \hline
\multicolumn{5}{c}{}\\
\end{tabular}
\label{table:NRItabEvt2}
\end{subtable}

\begin{subtable}[b]{1.0\textwidth}
\centering
\caption{cNRI computation based on the reclassification tables.}
\begin{tabular}{l|c}\hline
{\bf Censoring-adjusted NRI calculation}                                                       & {\bf Quantity} \\ \hline
Number of patients without events predicted in a lower risk category by Bayes-AC  ($\bar{E}_{M_1}^{*,\downarrow}$)& 1,753 \\
Number of patients without events predicted in a lower risk category by COX     ($\bar{E}_{M_2}^{*,\downarrow}$) & 1,914 \\
Total number of patients without events         ($n^*_{\bar E}$)                                                  & 37,624    \\
Number of patients with events predicted in a higher risk category by Bayes-AC ($E_{M_1}^{*,\uparrow}$)&  202 \\
Number of patients with events predicted in a higher risk category by COX     ($E_{M_2}^{*,\uparrow}$) &  102 \\
Total number of patients with events         ($n^*_E$)                                                  &  1,074  \\
&\\
Risk reclassification improvement for patients without events             & -0.0043 \\
Risk reclassification improvement for patients with events                 & 0.0931 \\ \hline
Total net risk reclassification improvement                            & 0.0888\\ \hline
\end{tabular}
\label{table:NRISummary}
\end{subtable}
\end{table}

\begin{figure}[ht]
\centering
\includegraphics[width=0.45\textwidth]{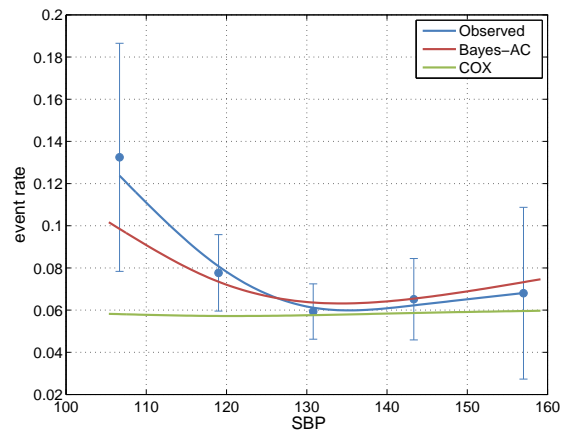}
\caption{Relationship between systolic blood pressure and CV risk for males who are on SBP medication. }
\label{figure:sbpVsriskMeds}
\end{figure}

Although the overall improvement in calibration was not very large (Figure~\ref{CalibCensBayesCox}), the Bayesian network model also allows for greater flexibility to model CV risk in certain subgroups.
For example, it is generally understood that the CV risk rises with increased blood pressure~\citep{stamler1993blood}. In most cases, our data supports this assertion. However for males who are already on a blood pressure medication, the relationship between the risk and blood pressure is not strictly increasing, as we see in Figure \ref{figure:sbpVsriskMeds}. The risk increases with decreasing SBP for SBP below 130 mmHg. This observation is interesting because physicians typically treat SBP down to 130 mmHg for people whose blood pressure is not controlled. Blood pressure being treated below 130 mmHg may indicate an underlying disease which is probably evident to a physician but is not captured by the risk factors that we have trained our model on. The underlying disease increases the risk for this group of patients.

The proportional hazards model that we are using forces a linear relationship between the log hazard and the log of SBP and, thus, ends up modeling the risk well only for people whose blood pressure is relatively elevated. This is evident in Figure \ref{figure:sbpVsriskMeds} which shows the Bayes-AC model with better calibration. In addition, the C-index of the Bayes-AC model (0.646) is significantly higher that of the COX model (0.576) for predictions in this subgroup.

\section{Discussion and conclusion}

This paper focuses on the application of a machine learning approach to risk prediction using EHD, when event times may be censored due to unequal individual follow-up. Traditional statistical models for event-time data with censored observations are well-developed, but typically less flexible than established machine learning techniques for classification. On the other hand, most classifiers do not handle censoring, as they assume that labels in the training data are fully observed (or in the case of a semi-supervised classifier, observed on a random subsample of individuals). Our proposed technique combines features from both of these approaches, using inverse probability weighting to extend the Bayesian network technique for censored event data. Although we apply our approach to a Bayesian network, IPCW can be extended to other machine learning classifiers.

In addition to offering both modeling flexibility and statistical validity, our technique seamlessly handles missing data, which is common in EHD, and offers the opportunity to combine findings from the medical literature with clinical judgment to shape the model. The advantages of our method are illustrated by applying it to a large electronic health database. We show that: (1) ignoring censoring when performing classification results in grossly mis-calibrated predictions, (2) the Bayesian network learns non-linear predictor-outcome relationships better than standard proportional hazards regression models of the type typically used to construct cardiovascular risk models from longitudinal studies with censored event data.

In addition to presenting a novel method, we also emphasize the importance of using appropriate criteria for assessing the performance of predictive models for censored data. Commonly-employed metrics for calibration and discrimination require event indicators which are fully known, and can be misleading in the presence of censoring. We present alternate metrics which are tailored to the censored data setting.

As we noted previously, the use of inverse probability of censoring weighting relies on the assumption that the censoring time is independent of the CV event time. Heuristically, we assume that patients more or less likely to have a CV event are not more likely to disenroll from the health system. We could relax this assumption by modeling the censoring time as a function of the risk factors.

Though there are known techniques for searching across multiple DAG structures in the context of Bayesian networks, we chose to focus on techniques for learning parameters for a given, fixed network. This decision was motivated by the fact that our clinical collaborators desire model interpretability and face validity, which may not be achieved with an automated process determining network topology. Further, flexible methods such as Bayesian networks are prone to overfitting. We control overfitting in our method by implementing BIC model averaging and bootstrapping; in our experiments, the method was not highly sensitive to tuning parameter values which determined the maximum number of mixture components in each model and the number of bootstrap resamples, provided these were set within reasonable ranges ($> 3$ and $>10$, respectively).

Though motivated by an example in electronic health data, our technique is applicable to any situation where event outcomes are subject to censoring. For example, in economics, one might wish to predict whether recently-unemployed individuals will be re-hired within a fixed time period, an outcome which is likely to be censored in most feasible study designs. In our context, we plan to incorporate this technique into a point-of-care clinical decision support system, which will provide more accurate cardiovascular risk predictions for patients based on their individual health history.

\section*{Acknowledgements}
This work was partially supported by NHLBI grant R01HL102144-01 and AHRQ grant R21HS017622-01.

\bibliographystyle{plainnat}
\bibliography{DMKD_140331_references_SB}
\end{document}